\documentclass[sigconf]{aamas}  
\usepackage{booktabs}

\usepackage{times}
\usepackage{xcolor}
\usepackage{soul}
\usepackage[utf8]{inputenc}

\usepackage{amssymb}
\usepackage{amsmath}
\usepackage{comment}
\usepackage{algorithm}
\usepackage{algorithmic}
\usepackage{amsmath}
\usepackage{moresize}
\usepackage{helvet}
\usepackage{courier}
\usepackage{graphicx}
\usepackage{wrapfig}
\usepackage{url}
\usepackage{subcaption}

\setcopyright{ifaamas}  
\acmDOI{doi}  
\acmISBN{}  
\acmConference[AAMAS'19]{Proc.\@ of the 18th International Conference on Autonomous Agents and Multiagent Systems (AAMAS 2019), N.~Agmon, M.~E.~Taylor, E.~Elkind, M.~Veloso (eds.)}{May 2019}{Montreal, Canada}  
\acmYear{2019}  
\copyrightyear{2019}  
\acmPrice{}  

\def\i#1{\hbox{\it #1\/}}
\def\beq{\begin{equation}}
\def\eeq#1{\label{#1}\end{equation}}
\def\ba{\begin{array}}
\def\ea{\end{array}}

\def\t{\textbf{t}}
\def\f{\textbf{f}}

\def\iif{\hbox{\bf if}}

\def\causes{\hbox{\bf causes}}
\def\inertial{\hbox{\bf inertial}}
\def\default{\hbox{\bf default}}

\def\nonex{\hbox{\bf nonexecutable}}

\begin{document}

\title[Integrating Task-Motion Planning with Reinforcement Learning]{Integrating Task-Motion Planning with Reinforcement Learning for Robust Decision Making in Mobile Robots
}

\author{Yuqian Jiang}
\affiliation{%
  \institution{Department of Computer Science, UT-Austin}
  \city{Austin} 
  \state{TX} 
}
\email{jiangyuqian@utexas.edu}
\author{Fangkai Yang}
\affiliation{%
  \institution{Maana Inc.}
  \city{Bellevue} 
  \state{WA} 
}
\email{fyang@maana.io}
\author{Shiqi Zhang}
\affiliation{%
  \institution{Department of Computer Science, SUNY-Binghamton}
  \city{Binghamton} 
  \state{NY} 
}
\email{szhang@cs.binghamton.edu}
\author{Peter Stone}
\affiliation{%
  \institution{Department of Computer Science, UT-Austin}
  \city{Austin} 
  \state{TX} 
}
\email{pstone@cs.utexas.edu}

\begin{abstract}
Task-motion planning (TMP) addresses the problem of efficiently generating executable and low-cost task plans in a discrete space such that the (initially unknown) action costs are determined by motion plans in a corresponding continuous space. However, a task-motion plan can be sensitive to unexpected domain uncertainty and changes, leading to suboptimal behaviors or execution failures. In this paper, we propose a novel framework, \emph{TMP-RL}, which is an integration of TMP and reinforcement learning (RL) from the execution experience, to solve the problem of robust task-motion planning in dynamic and uncertain domains. TMP-RL features two nested planning-learning loops. In the inner TMP loop, the robot generates a low-cost, feasible task-motion plan by iteratively planning in the discrete space and updating relevant action costs evaluated by the motion planner in continuous space. In the outer loop, the plan is executed, and the robot learns from the execution experience via model-free RL, to further improve its task-motion plans. RL in the outer loop is more accurate to the current domain but also more expensive, and using less costly task and motion planning leads to a jump-start for learning in the real world. Our approach is evaluated on a mobile service robot conducting navigation tasks in an office area. Results show that TMP-RL approach significantly improves adaptability and robustness (in comparison to TMP methods) and leads to rapid convergence (in comparison to task planning (TP)-RL methods). We also show that TMP-RL can reuse learned values to smoothly adapt to new scenarios during long-term deployments. 

\end{abstract}


\maketitle

\section{INTRODUCTION}

Building mobile robots that behave intelligently in real environments is one of the central problems of robotics and artificial intelligence. In many practical scenarios, the robot is given a request from an external human user, such as ``deliver coffee to Alice''. To achieve a goal like this, the robot needs to perform {\em motion planning} (MP) through continuous space \cite{choset2005principles} using algorithms such as Probabilistic Random Map \cite{kayraki1996probabilistic} or Rapidly-exploring Random Trees \cite{lavalle1998rapidly} to generate a collision-free trajectory based on the current status of the environment, and complete a long sequence of navigation and manipulation actions. Generating a long-horizon motion plan is computationally expensive and not worthwhile, because it can quickly become out-of-date in uncertain and dynamic domains. To mitigate this problem, a higher-level planning layer, called {\em task planning} (TP), is introduced \cite{ghallab2004automated}. Task planning is performed on a more abstract representation of the dynamic environment, by modeling segments of motion trajectory in continuous spaces as a sequence of high-level atomic transitions, such as moving from A to B or picking up coffee from the table, in a discrete space such that each step of the task plan can be expanded into a corresponding motion plan in the original continuous space. Leveraging decades of research on classical AI planning \cite{cim08}, common approaches to represent domain dynamics in terms of high-level subtasks include using PDDL \cite{mcdermott1998pddl} or an action language \cite{gel98} that relates to logic programming under answer set semantics (answer set programming) \cite{lif08}. Given a domain representation, a task planner, e.g., a PDDL-based solver such as \textsc{FastDownward} \cite{helmert2006fast} or an answer set solver such as \textsc{Clingo} \cite{gekasc12c}, can generate a sequence of subtasks, and a motion planner is used to check the feasibility and/or cost of each subtask, to ensure the task plan generated from discrete domain is executable and efficient in continuous domain. The integration between high-level task planning and low-level motion planning, also known as {\em task-motion planning} (TMP), has been widely studied for manipulation tasks \cite{erdem2011combining,srivastava2014combined,garrett2018ffrob} and navigation tasks \cite{chen2015task,lo2018petlon}, reducing the complexity of long-horizon motion planning and improving plan feasibility, quality and scalability.

Despite the progress made on generating feasible and quality \textit{plans} based on prior discrete and continuous modeling of the domain, during \textit{execution} of the planned actions in the real world, a robot can still face domain uncertainties and changes that are not available at modeling time. For instance, motion planning for navigation may not model congested paths that keep changing during the day or direct sunlight that reduces success rate of navigating through an area. Consequently, such unexpected changes may invalidate task-motion plans, leading to suboptimal behaviors and execution failures. Continually learning from execution experience and adapting to the changing domain is a prerequisite for long-term autonomy \cite{biswas20161,khandelwal2017bwibots}. To this end, reinforcement learning (RL) \cite{sutton1998reinforcement} has been used to build highly-adaptive autonomous agents \cite{mnih2015human}. However, many RL algorithms require relatively large amounts of data, which is expensive or sometimes dangerous to obtain for real robots. Recent work has focused on leveraging symbolic planning to guide RL \cite{leonetti2016synthesis,yang:peorl:2018,yang:sdrl:2018} by accelerating learning and improving sample efficiency. Despite their success in simulation environments, for applications in real robots, the learning expense is still quite high, because these approaches do not leverage a motion planner to provide a cheaper first-step evaluation on the plan quality and feasibility before sending a potentially infeasible or costly plan for execution and learning. In the TMP-RL approach introduced in this paper, task plans are first evaluated by motion planners to ensure their quality and feasibility, before being proposed to execution and learning.

In order to {\em improve adaptability and robustness of task-motion plans for real robots}, in this paper, we propose to integrate TMP with RL such that the robot can constantly generate feasible, high-quality task-motion plans and rapidly learn from execution experience to adapt to domain changes. Inspired by PETLON, a recent task-level-optimal TMP algorithm~\cite{lo2018petlon}, and PEORL, a state-of-the-art task planning-RL architecture \cite{yang:peorl:2018}, our approach features {\em two nested planning--reinforcement learning loops}:
\begin{itemize}
\item The {\em inner loop} is a complete TMP algorithm, where a symbolic plan is generated and each symbolic action is evaluated by the motion planner. 
Value iteration is performed on rewards derived from motion plan costs, and the learned values are sent back to task planner to generate an improved task plan in the next evaluation episodes. By the end of the iterative loop, a feasible and high-quality task plan conditioned on motion plan costs is generated. 
\item The {\em outer loop} is for learning to generate an optimal task-motion plan to accommodate domain uncertainty, change, and extra reward information. A task-motion plan generated by the inner loop is sent for execution, and value iteration is performed on rewards derived from real execution experience including navigation cost, execution failure, and environmental reward. After one episode of TMP, execution, and learning, the learned values are sent to the inner loop, so as to generate an improved task-motion plan for the next episode. When the outer loop terminates, the robot has learned a task-motion plan that has adapted to the observed domain changes.
\end{itemize}
With the architecture above, the inner loop gives the robot sufficient deliberation to generate a good quality task-motion plan based on its own discrete and continuous models, leads to a jump-start of plan quality, and reduces the chance of sending infeasible or known sub-optimal plans for execution and learning directly from the environment, which may be expensive and sometimes dangerous. The outer-loop further fine-tunes the task-motion plans by learning from the environment, improving the robustness and adaptability of TMP facing domain uncertainty and change. The duality between the inner loop and outer loop allows a seamless integration of TMP with RL such that motion planning in continuous model and reinforcement learning from the real execution experience can jointly contribute to improving TMP. Due to the jump-start of the quality of task-motion plans, the learning efficiency can be further improved.

Our approach is generic in the sense that a variety of task planning, motion planning, and reinforcement learning approaches can be used.  
In this paper, we instantiate our approach using the same symbolic planning and reinforcement learning technology as PEORL \cite{yang:peorl:2018}, including action language $\mathcal{BC}$ \cite{lee2013action} for task planning due to its expressiveness, formal semantics and efficient implementation using answer set solver {\sc Clingo}, and R-learning \cite{sm:mlj96} for reinforcement learning. R-learning is an important
family of reinforcement learning paradigm that characterizes
finite horizon average reward, and is shown to be particularly
suitable for planning and scheduling tasks.
We evaluate the approach using the Gazebo simulator~\cite{koenig2004design} for a real service robot platform and the environment it operates in. Compared to PETLON, a recent task-level-optimal TMP algorithm~\cite{lo2018petlon} and PEORL, a recent Task Planning (TP)-RL approach~\cite{yang:peorl:2018}, TMP-RL demonstrates superior adaptability to environmental uncertainties and achieves better task performance than PETLON, and more efficient exploration and faster convergence than PEORL. The experiment is extended with a sequence of different scenarios, showing that TMP-RL can smoothly reuse learned information to improve long-term performances.


\section{RELATED WORK}

Task planning~\cite{ghallab2004automated} and motion planning~\cite{choset2005principles} algorithms generate plans in symbolic and continuous spaces respectively. 
Although robots that operate in the real world need capabilities of both task and motion planning, it is not until recent years that the term of Task and Motion Planning (TMP) was used in the literature to refer to algorithms that integrate both planning paradigms~\cite{dantam2018incremental,srivastava2014combined,garrett2018ffrob,lo2018petlon,kaelbling2013integrated,lagriffoul2014efficiently,chen2015task}. 
These TMP algorithms have different focuses, such as ensuring symbolic actions' feasibility via motion planning~\cite{srivastava2014combined}, integrated symbolic planning under uncertainty and motion planning~\cite{kaelbling2013integrated}, and leveraging symbolic search heuristics in motion planning space~\cite{chen2015task,garrett2018ffrob}. 
Among the TMP algorithms, PETLON~\cite{lo2018petlon}, which uses sampling-based probabilistic motion planning methods to evaluate costs of task-level actions is most similar to the inner loop of our work, but in our work, we use RL to learn rewards derived from real action costs, whereas PETLON is purely a planning method: it assumes that the resulting task-motion plan can be executed with no further changes. While generating feasible, low-cost task-motion plans is the major focus of existing work on TMP, to the best of our knowledge, mixing task-motion planning and learning from execution to accommodate domain uncertainty and change has not been investigated before.

The integration of symbolic planning with reinforcement learning has been studied in a variety of approaches \cite{parr1998reinforcement,Ryan02usingabstract,hogg2010learning,leonetti2016synthesis}. These methods focus on leveraging the strengths of one of the paradigms to enhance the other. Integrating robot task planning and learning of navigation costs has also been investigated \cite{kha14}. Recent approaches such as PEORL \cite{yang:peorl:2018} and SDRL \cite{yang:sdrl:2018} utilize closed-loop communication between planning and learning: an optimal symbolic plan is obtained from an iterative process of planning and learning, so that planning and learning can mutually benefit each other. However, most of these approaches have only been applied to artificial domains. For instance, in Atari games, an avatar can die numerous times before learning not to jump off the cliff. Learning on real robots to perform trial-and-error of this kind can be quite expensive or even dangerous. On the other hand, an integrated robot system is usually equipped with pre-mapped environment landscape and motion planners that can be used to evaluate the outcomes of task plans before executing the plan and learning from the environment, calling for an integration of TMP with RL. Our approach is inspired by PEORL and PETLON, but generalizes them into two nested loops to capture the complete task-motion planning, execution and learning loop for mobile robots. The two nested loops allow estimates made by motion planner and values learned from the environment to jointly improve the quality of plans. Consequently, TMP-RL is adaptive to real-world changes like PEORL while efficiently leveraging a motion planner to generate economical task plans like PETLON. To the best of our knowledge, this is the first work to apply reinforcement learning to improve robustness and adaptability of task-motion plans, where task planning in discrete spaces and motion planning, execution and learning in continuous spaces are handled in a unified framework.

\section{Preliminaries}
In this section, we individually introduce the symbolic planning, motion planning and learning technologies that will be combined in our framework introduced in Section~\ref{sec:framework}.
\subsection{Symbolic Planning}
\label{sec:symbolic}
An {\em action description} $\mathbb{D}$ in the language $\mathcal{BC}$ \cite{lee13} includes two kinds of symbols, {\em fluent constants} that represent the properties of the world, denoted as $\sigma_F(\mathbb{D})$, and {\em action constants}, denoted as $\sigma_A(\mathbb{D})$. A fluent atom is an expression of the form $f=v$, where $f$ is a fluent constant and $v$ is an element of its domain. For boolean domain, denote $f=\t$ as $f$ and~$f=\f$ as $\sim\!\!\! f$. An action description is a finite set of {\em causal laws} that describe how fluent atoms are related with each other in a single time step, or how their values are changed from one step to another, possibly by executing actions. For instance,
$$
A~\iif~A_1,\ldots,A_m
$$
is a {\em static law} that states at a time step, if $A_1,\ldots, A_m$ holds then $A$ is true. Another static law
$$
\default~f=v
$$
states that by default, the value of $f$ equals~$v$ at any time step. 
$$
a~\causes~A_0~\iif~A_1,\ldots, A_m
$$
is a {\em dynamic law}, stating that at any time step, if $A_1,\ldots, A_m$ holds, by executing action $a$,~$A_0$ holds in the next step. 
$$
\nonex~a~\iif~A_1,\ldots,A_m
$$
states that at any step, if $A_1,\ldots, A_m$ holds, action $a$ is not executable. Finally, the dynamic law
$$
\inertial~f
$$
states that by default, the value of fluent $f$ does not change from one step to another, formalizing the {\em commonsense law of inertia} that addresses the frame problem.

An action description captures a dynamic transition system. A {\em state} $s$ is a complete set of fluent atoms, and a transition is a tuple $\langle s_1,a,s_2 \rangle$ where $s_1, s_2$ are states and~$a$ is a (possibly empty) set of actions. The semantics of $D$ is defined by a translation into a set of answer set programs~$\i{PN}_l(\mathbb{D})$, for an integer~$l\ge 0$ stating the maximal steps of transition. It is shown that all answer sets of $\i{PN}_0(\mathbb{D})$ correspond to all states in the transition system, and all answer sets of $\i{PN}_l(\mathbb{D})$ correspond to all transition paths $\Pi$ of length $l$, of the form $\langle s_1,a_1,\ldots, a_{l-1}, s_l \rangle$ (or equivalently, $\Pi=\bigcup^{l-1}_1\langle s_i, a_i, s_{i+1}\rangle$) \cite[Theorems 1, 2]{lee13}. Let $\mathbb{I}$ and $\mathbb{G}$ be states. The triple $(\mathbb{I},\mathbb{G},\mathbb{D})$ is called a planning problem. $(\mathbb{I},\mathbb{G},\mathbb{D})$ has a plan of length $l-1$ iff there exists a transition path of length $l$ such that $\mathbb{I}=s_1$ and $\mathbb{G}=s_l$, which is encoded in the answer set of $\bigcup_{i=1}^l \i{PN}_l(\mathbb{D})$. Throughout the paper, we use $\Pi$ to denote both the plan and the transition path by following the plan.  Automated planning can be achieved by an answer set solver.

\subsection{Motion Planning}
Motion planning is one of the most important research areas in robotics, and aims at planning in continuous spaces to connect a start configuration $S$ and a goal configuration $G$ while avoiding collisions with obstacles~\cite{choset2005principles}. 
Robot motion is frequently represented as a path in a configuration space, which is potentially higher-dimensional.
A configuration space includes a set of all possible configurations, where a configuration describes a possible pose of the robot. 
The output of a motion planner includes a sequence of discrete motions that can be directly passed to the joints of robot for execution. 
In this work, we consider a mobile robot that moves in 2D spaces. where 
we directly search in the 2D workspace (instead of higher-dimensional configuration space). 
A motion planning problem can be specified by an
initial position $x^{init}$ and a goal set $X^{goal}$. 
The 2D space is represented as a region in Cartesian space such that the position and
orientation of the robot can be uniquely represented as a \emph{pose} $({\bf
x},\theta)$. Some parts of the space are designated as free space, and the rest
is designated as obstacle. 

The motion planning problem is solved by the motion planner $\mathcal{P}^{geo}$ to compute a
collision-free trajectory $\xi^*$ (connecting ${\bf x}^{init}$ and a pose ${\bf
x}^{goal}\in {\bf X}^{goal}$ taking into account any motion constraints on the part of the robot)
with minimal trajectory length $Len(\xi)=L$.
We use $\Xi$ to represent the
trajectory set that includes all satisfactory trajectories. The \emph{optimal}
trajectory is 
\begin{align*}
    \xi^* = \text{argmin}_{\xi \in \Xi} Len(\xi), 
\end{align*}
where $\xi(0) = {\bf x}_{init}$ and $\xi(L) = {\bf x}_{goal}\in {\bf X}_{goal}$. 
In particular, we use \texttt{\small global\_planner}, an off-the-shelf package from the Robot Operating System (ROS)~\cite{quigley2009ros} community for motion planning, which generates trajectories using gradient descent method together with standard $A^*$ and Dijkstra's search.

\subsection{R-learning for Finite Horizon Problems.} 

A Markov Decision Process (MDP)   is defined as a tuple $({\mathcal{S},\mathcal{A},P_{ss'}^{a},r,\gamma})$, where $\mathcal{S}$ and $\mathcal{A}$ are the sets of symbols denoting states and actions, the transition kernel $P_{ss'}^{a}$ specifies the probability of transition from state $s\in\mathcal{S}$ to state $s'\in\mathcal{S}$ by taking action $a\in\mathcal{A}$, $r(s,a):\mathcal{S}\times\mathcal{A}\mapsto\mathbb{R}$ is a reward function bounded by $r_{\max}$, and $0\leq\gamma<1$ is a discount factor. A solution to an MDP is a policy $\pi:\mathcal{S}\mapsto \mathcal{A}$ that maps a state to an action. RL concerns on learning a near-optimal policy  by executing actions and observing the state transitions and rewards, and it can be applied even when the underlying MDP is not explicitly given, a.k.a, model-free policy learning.

To evaluate a policy $\pi$, there are two types of performance measures: the expected discounted sum of reward for infinite horizon problems 
and the expected un-discounted sum of reward for finite horizon problems. In this paper we adopt the latter metric. Define $J^\pi_{\rm avg}(s) = \mathbb{E}[\sum\limits_{t = 0}^T {{r_t}}|s_0=s ]$, and the \textit{gain reward} ${\rho ^\pi }(s)$ reaped by policy $\pi$ from $s$ as
$$
{\small
{\rho ^\pi }(s) = \mathop {\lim }\limits_{T \to \infty } \frac{{J^\pi_{{\rm{avg}}}(s)}}{T} = \mathop {\lim }\limits_{T \to \infty } \frac{1}{T}\mathbb{E}[\sum\limits_{t = 0}^T {{r_t}} ]
} .
$$ 
R-learning \cite{rlearning:schwartz,sm:mlj96} 
 is a model-free value iteration algorithm that can be used to find the optimal policy for the average reward criterion. At the $t$-th iteration $(s_t,a_t,r_t, s_{t+1})$, the following update is performed:
\beq
\ba{rl}
 {R_{t + 1}}({s_t},{a_t})\!\! & \xleftarrow{\alpha_t} {r_t} - {\rho _t}(s_t)  
 + \mathop {\max }\limits_a {R_t}({s_{t + 1}},a)\\
 \rho_{t+1}(s_t)\!\!& \xleftarrow{\beta_t} r_t + \mathop {\max }\limits_a {R_t}({s_{t + 1}},a) - \mathop {\max }\limits_a {R_t}({s_t},a),
\ea
\eeq{riter1}
where $\alpha_t, \beta_t$ are the learning rates, and $a_{t+1} \xleftarrow{\alpha} b$ denotes the soft update rule ${a_{t + 1}} = (1-\alpha){a_{t}} + \alpha b$.


\section{TMP-RL Framework}
\label{sec:framework}
\begin{figure}
\centering
\includegraphics[height=5cm,width=8cm]{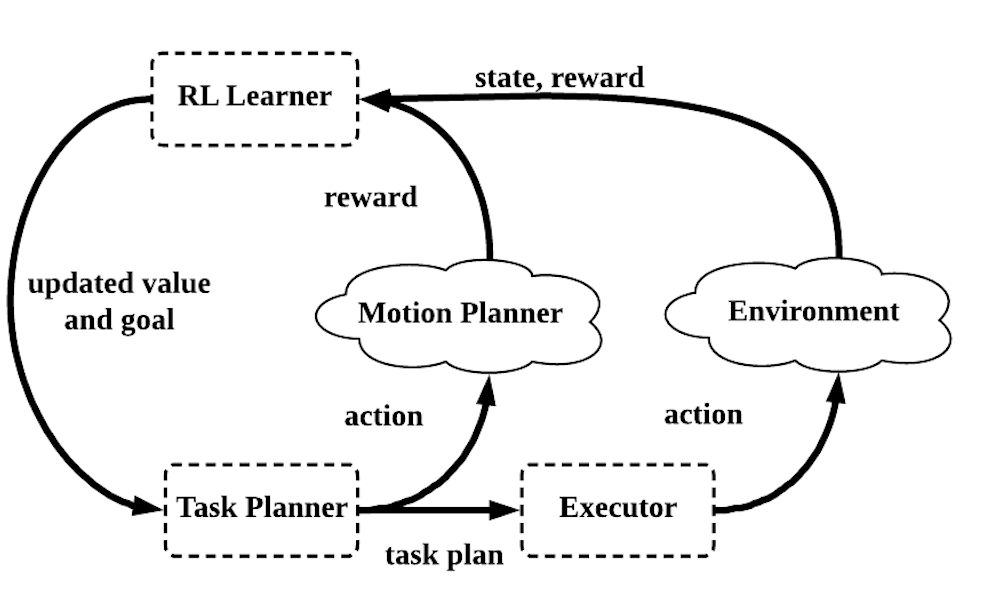}
\caption{An illustration of our TMP-RL framework}
\label{fig:arch}
\end{figure}
The TMP-RL framework we propose is shown in Fig.~\ref{fig:arch}. The inner loop consists of a task planner, a motion planner and a reinforcement learner that iteratively performs planning and learning to generate {\em a feasible and low-cost task-motion plan}. Once the inner loop returns a task-motion plan, it is sent to execution in the outer loop, where the reinforcement learner performs value iteration on the reward derived from execution experience. The learned values are returned into the symbolic planner, and the inner loop runs again to generate an improved task-motion plan leveraging motion planner and learned experience. The architecture is explained in detail below.

\subsection{Optimal Task Plan Conditioned on Motion Plan}


A task planning problem defines the objective of generating a satisfactory plan $\Pi^\tau$, i.e., a sequence of actions given a planning problem $(\mathbb{I}^\tau,\mathbb{G}^\tau,\mathbb{D}^\tau)$, where $\mathbb{D}^\tau$ is a domain independent symbolic formulation given by human expert, $\mathbb{I}^\tau$ is an initial state and $\mathbb{G}^\tau$ a goal state. As in PEORL, $\mathbb{D}^\tau$ consists of causal laws that formulates precondition and effects of actions, such as $approach$ door $D_1$ causes the robot besides $D_1$ if currently the robot is beside $D_2$ and $D_1$ is accessible from $D_2$:
$$
\i{approach}(D_1)~\causes~\i{besides}(D_1)~\iif~\i{beside}(D_2),\i{acc}(D_2,D_1)
$$
and static relationship on fluents, such as symmetry of accessible relationship:
$$
\i{acc}(D_1,D_2)~\iif~\i{acc}(D_2,D_1)
$$

A motion planning problem concerns on generating a collision free trajectory $\xi(\mathbb{I}^m,\mathbb{G}^m)$ given a motion planning problem $(\mathbb{D}^m, \mathbb{I}^m,\\ \mathbb{G}^m)$ where $\mathbb{D}^m$ is a motion planning domain, $\mathbb{I}^m$ is an initial position and $\mathbb{G}^m$ is the goal position, such that the position $\mathbb{I}^m$ is connected with position $\mathbb{G}^m$.

We use a mapping function $f:X=f(s)$ that maps a symbolic state $s$ into a set of feasible poses $X$ in continuous space, for the motion planning algorithm to sample from. We assume the availability of at least one pose $x\in X$ in each state $s$, such that the robot is in the free space of $\mathbb{D}^m$. If it is not the case, the state $s$ is declared infeasible. Given a motion planning domain $\mathbb{D}^m$ and a task plan $\Pi^\tau$ for task planning problem $(\mathbb{D}^\tau, \mathbb{I}^\tau, \mathbb{G}^\tau)$, a plan refinement of $\Pi^\tau$ w.r.t motion planner, denoted as $\Pi^m$, is a sequence of collision free trajectories obtained by perform motion planning on each navigation actions, i.e.,
\beq
\Pi^m = \bigcup_{\langle s,a,s'\rangle\in\Pi^\tau}\xi(x,x')
\eeq{refinement}
where $x\in f(s)$, $x'\in f(s')$.

The cumulative cost of a task plan $\Pi^\tau$ is obtained by the cumulative length of its motion planning refinement $\Pi^m$, i.e., 
$
\i{Cost}(\Pi^\tau)=\sum_{\xi\in \Pi^m}\i{Len}(\xi).
$
An {\em optimal task plan conditioned on motion plan} is defined the task plan $\Pi^\tau_o$ such that $\Pi^m_o$ has the minimal length among all task plans.

Although motion planning is typically not as expensive as RL methods that learn from the real world, evaluating costs and feasibilities of many symbolic actions is still very time-consuming. Even though motion plans for all pair-wise positions mapped from symbolic states are pre-computed, computing the optimal symbolic plan $\Pi^\tau_o$ in general is still PSPACE-complete. In order to iteratively approximate $\Pi^\tau_o$, we integrate task planning and motion planning using R-learning. 
\subsection{Task Motion Planning with Reinforcement Learning}
\subsubsection{Reward}
Given a symbolic transition $\langle s,a,s'\rangle$ where $a$ can be refined by motion planner, we define a reward function $r$ that is negative and inversely proportional to a distance metric of the motion plan that refines the navigation action~$a$, mapped by a function $R:\mathbb{R}^+\mapsto\mathbb{R}^-$:
$$
r(s,a)=R(\i{Len}(\xi(x,x'))\propto \frac{1}{\i{Len}(\xi(x,x'))},
$$
where $x\in f(s), x'\in f(s')$. One simple way to instantiating~$R$ is 
$$r(s,a) = R(Len(\xi(x,x')))=-Len(\xi(x,x')).$$

If motion plan fails for transition $\langle s,a,s'\rangle$, define $r(s,a)=-\infty$.

\subsubsection{Domain Formulation}
\label{sec:domain}
We enrich the domain formulation $\mathbb{D}^\tau$ with the following causal laws formulating the effect of actions on cumulative plan quality:
$$
a~\causes~\i{quality}=C+Z~\iif~s,\rho(s,a)=Z,\i{quality}=C
$$
where $s$ is a state. The $\rho$-values are initialized optimistically to the upper-bound of gain reward, which is the reward derived from the $L_p$ metric in the configuration space:
$$
\default~\rho(s,a)=\max_{x,x'} R(||x-x'||_p)
$$
where $x\in f(s),x'\in f(s'),x\neq x'$, for $\langle s,a,s'\rangle$ in $T(\mathbb{D}^\tau)$, $p\in\mathbb{R}^+$, and $L_p$ metric stands for
$$
||x-x'||_p = \left(\sum_{i=1}^n|x_i-x'_i|^p\right)^{-p}.
$$
\subsubsection{Planning Goal}
At any episode $t$, planning goal $\mathbb{G}^\tau_t$ is contains a regular logical constraint describing the goal condition plus a linear constraint of the form 
\beq
\i{quality}\ge \i{quality}(\Pi^\tau_t)
\eeq{quality}
where $$\i{quality}(\Pi^\tau_t)=\sum_{\langle s,a,s'\rangle\in\Pi^\tau_t} \rho(s,a)$$ for some task plan~$\Pi^\tau_t$. In the planning -- learning loop, the linear constraint guides the planner to generate a plan with cumulative quality higher than a previous one, measured by learned $\rho$-values, leading to the iterative process of plan improvement based on reinforcement learning.
\subsubsection{Algorithm for TMP}
Algorithm~\ref{algexec} describes our inner loop of task-motion planning. 
The input to the algorithm includes a motion planning domain and a task planning problem. $q_0$ is initialized to be $-\infty$, and $P_0=\emptyset$. The algorithm first generates a task plan (Line 4). Then it iterates on each symbolic transition in the plan, and for each navigation action, it obtains the initial and goal poses in 2D domain (Line 12), generates motion plan (Line 13) and returns reward (Line 14). Value iteration of R-learning is performed with the reward (Line 15). At the end of this process, plan quality is computed using the learned $\rho$ values (Line 17), and it is used as the new constraint in the planning goal (Line 18), setting a baseline for the planner in the next iteration. The learned $\rho$ values are also updated in the symbolic formulation (Line~19). When the algorithm terminates, it outputs a task plan that cannot be further improved w.r.t motion planner.

\begin{algorithm}[htb!]
{\small
  \caption{Task-Motion Planning}
  \label{algexec}
  \begin{algorithmic}[1]
    \REQUIRE $(\mathbb{I}^\tau,\mathbb{G}^\tau,\mathbb{D}^\tau,f,\mathbb{D}^m, q_0, P_0)$ where $\i{quality}>q_0\in G^\tau$, and an exploration probability $\epsilon$
    \STATE $t\Leftarrow 0$
    \WHILE{$t< +\infty$}
      \STATE $\Pi^*\Leftarrow \Pi_t^\tau$
      \STATE obtain a plan $\Pi_t^\tau\Leftarrow~\i{Plan}(\mathbb{I}^\tau,\mathbb{G}^\tau,\mathbb{D}^\tau\cup P_t)$
      \IF {$\Pi^\tau_t=\emptyset$}
          \RETURN $\Pi^*$
      \ENDIF
      \FOR {symbolic transition $\langle s, a, s'\rangle\in\Pi^\tau_t$}
      \IF {$a$ cannot be refined by motion planner}
      	\STATE continue
      \ENDIF
      \STATE obtain initial pose $x=f(s)$ and goal pose $x'=f(s')$
  	  \STATE generate motion plan $\xi(x,x')$
      \STATE calculate reward $r(s,a)=R(\i{Len}(\xi(x,x')))$ 
      \STATE update $R(s,a)$ and $\rho^a(s)$ using (\ref{riter1}).
      \ENDFOR
      \STATE calculate quality of $\Pi^\tau_t$ by (\ref{quality}).
      \STATE update planning goal $G\Leftarrow (\i{quality}> \i{quality}_t(\Pi^\tau_t))$.
      \STATE update facts $P_t\Leftarrow \{\rho(s,a)=z:\langle s,a,s'\rangle\in\Pi, \rho_t^{a}(s)=z\}$
      \STATE $t\Leftarrow t+1$
    \ENDWHILE
  \end{algorithmic}}
\end{algorithm}

\begin{algorithm}[htb!]
{\small
  \caption{Task-Motion Planning and Learning}
  \label{algexec2}
  \begin{algorithmic}[1]
    \REQUIRE $(\mathbb{I}^\tau,\mathbb{G}^\tau,\mathbb{D}^\tau,f,\mathbb{D}^m)$ where $\i{quality}>0\in G^\tau$, and an exploration probability $\epsilon$
    \STATE $P_0\Leftarrow \emptyset$, $\Pi^\tau_0\Leftarrow \emptyset$, $q_0=-\infty$, $t=0$
    \WHILE{$t< +\infty$}
      \STATE $\Pi^*\Leftarrow \Pi^\tau_t$
      \STATE obtain a task-motion plan by calling Algorithm~\ref{algexec} $\Pi^\tau_t\Leftarrow\i{TMP}(\mathbb{I}^\tau,\mathbb{G}^\tau,\mathbb{D}^\tau,f,\mathbb{D}^\mu, q_t, P_t)$.
      \IF {$\Pi^\tau_t=\emptyset$}
          \RETURN $\Pi^*$
      \ENDIF
      \FOR {symbolic transition $\langle s, a, s'\rangle\in\Pi^\tau_t$} 
      \STATE execute $a$ and obtain true reward~$r(s,a)$.
      \STATE update $R(s,a)$ and $\rho^a(s)$ using (\ref{riter1}).
      \ENDFOR
      \STATE calculate quality of $\Pi^\tau_t$ by (\ref{quality}).
      \STATE update plan quality $q_t \Leftarrow \i{quality}_t(\Pi^\tau_t)$.
      \STATE update facts $P_t\Leftarrow \{\rho(s,a)=z:\langle s,a,s'\rangle\in\Pi^\tau_t, \rho_t^{a}(s)=z\}$
      \STATE $t\Leftarrow t+1$
    \ENDWHILE
  \end{algorithmic}}
\end{algorithm}

\subsection{TMP Execution and Learning}
Once a task-motion plan is generated, it is sent for execution, which goes to the outer loop of learning from real execution experience, where Algorithm~\ref{algexec} becomes the planning subroutine (Line 4) in Algorithm~\ref{algexec2}. In Algorithm~\ref{algexec2}, each action is executed in the environment (Line 12,15), and the true reward is obtained for  actions (Line 12). The value iteration performed on the true reward received during execution further rewrites the value learned through motion planner and feed back into the TMP algorithm (Line 20) to iteratively generate a task-motion plan that is adaptable to domain change.

It can be seen that Algorithm~\ref{algexec2} is very similar to Algorithm~\ref{algexec}, with the only difference being Line 4 in Algorithm~\ref{algexec} is a task planning call and Line 4 in Algorithm~\ref{algexec2} is a task-motion planning call. The duality between the inner loop and the outer loop brings a unification of refining task plans through motion planner and through learning from the environment: the quality of task plans are learned in the same way and the learned values are propagated back into symbolic representation so that  in the iterative learning process, motion planers and execution experience can jointly contribute to the plan improvement. 

\begin{figure*}[!hbt]
  \centering
  \begin{subfigure}{0.23\columnwidth}
  	\includegraphics[width=\columnwidth]{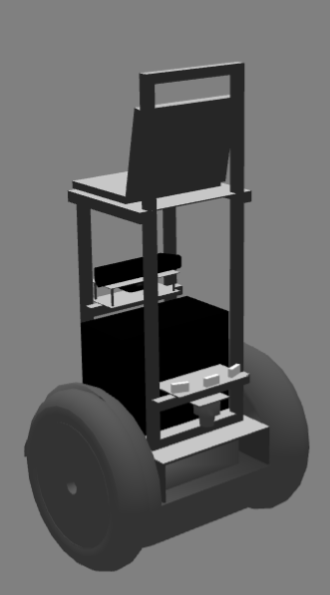}	
  	\caption{Simulated robot}	
    	\label{fig:bwibot}
  \end{subfigure}
  \quad
  \begin{subfigure}{0.67\columnwidth}
  	\includegraphics[width=\columnwidth]{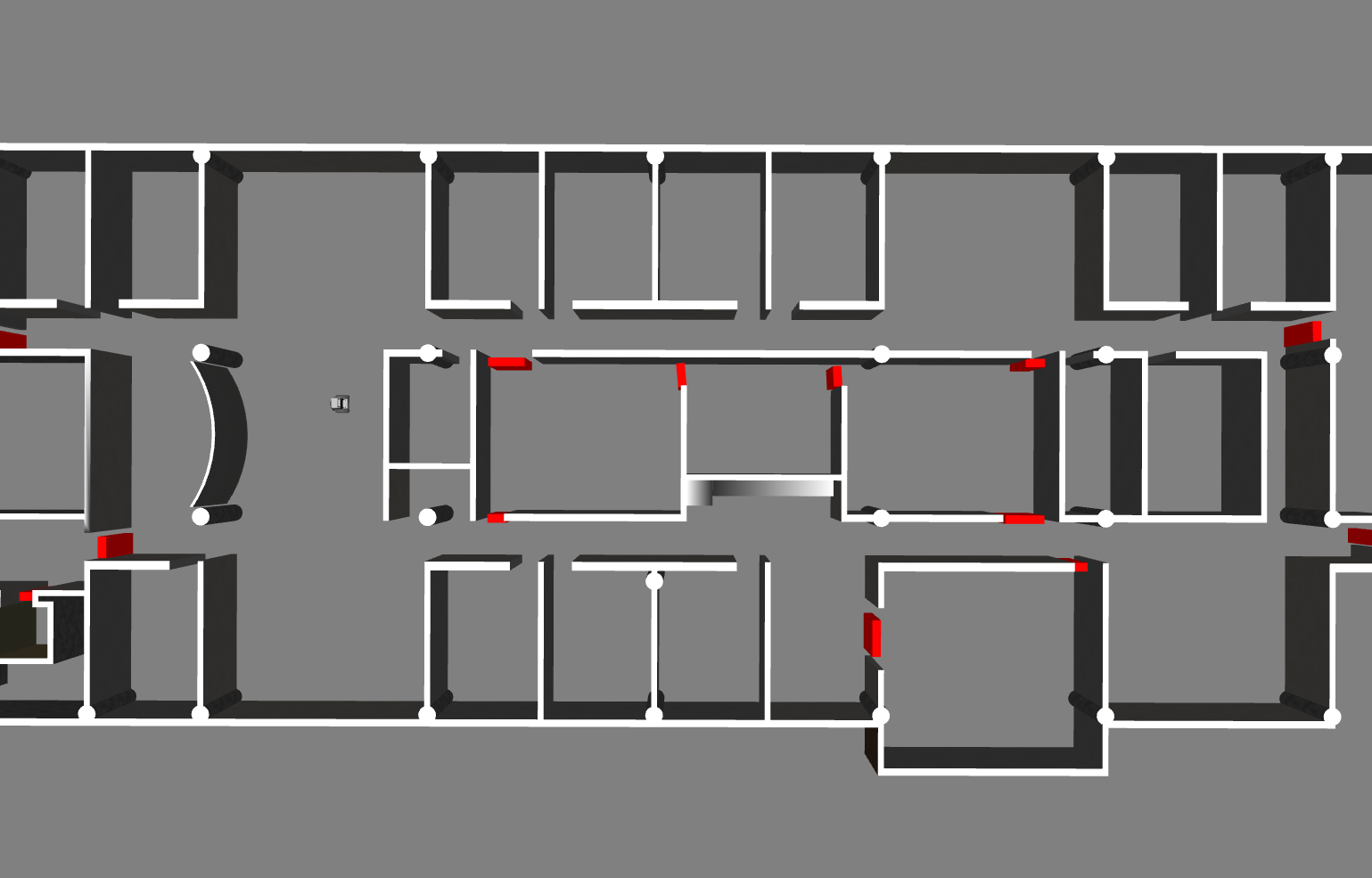}
  	\caption{Gazebo simulation environment}	
    	\label{fig:gazebo}
  \end{subfigure}
  \quad
  \begin{subfigure}{0.31\columnwidth}
  	\includegraphics[width=\columnwidth]{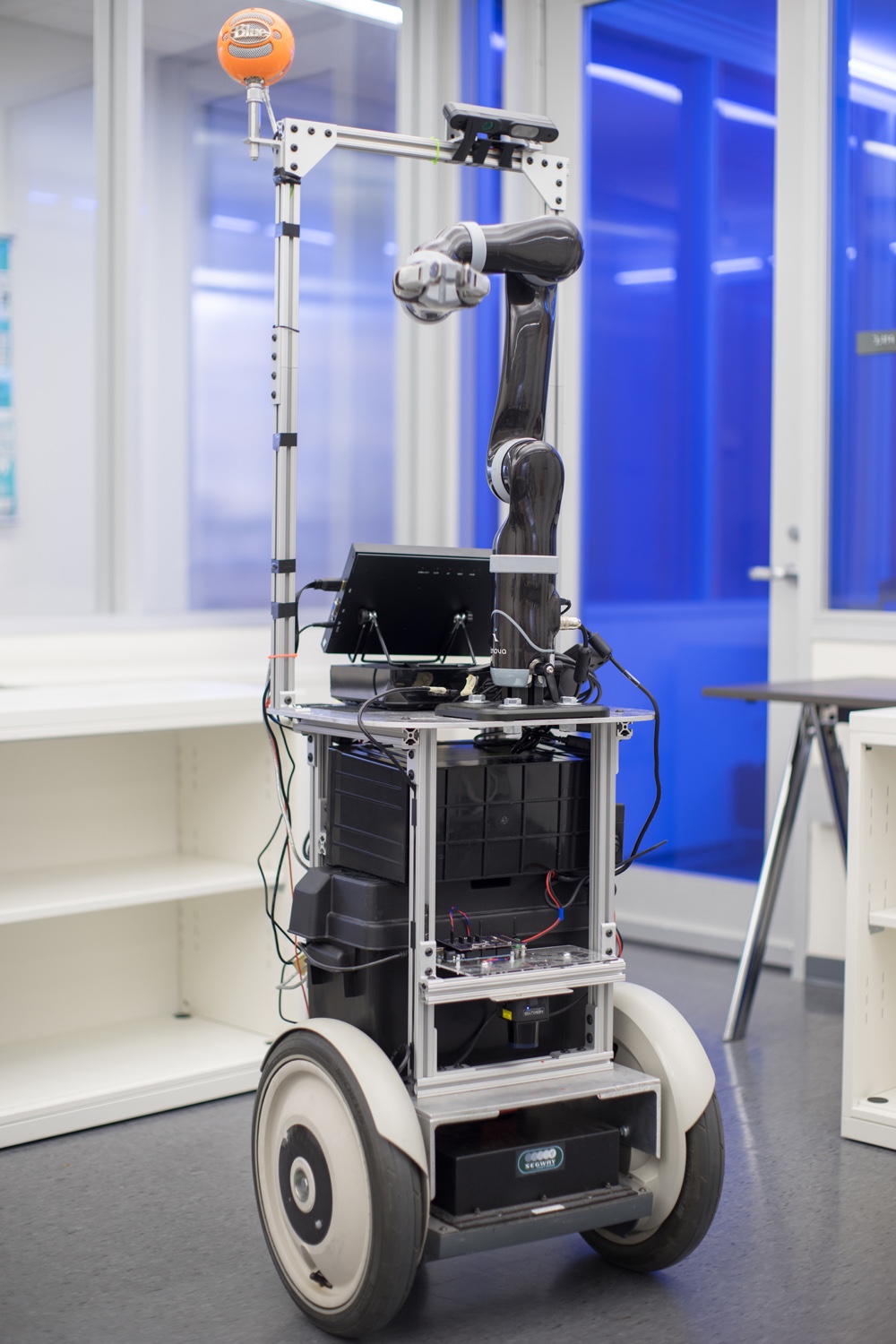}	
  	\caption{Real robot}	
    	\label{fig:bwibot_real}
  \end{subfigure}
  \quad
  \begin{subfigure}{0.65\columnwidth}
  	\includegraphics[width=\columnwidth]{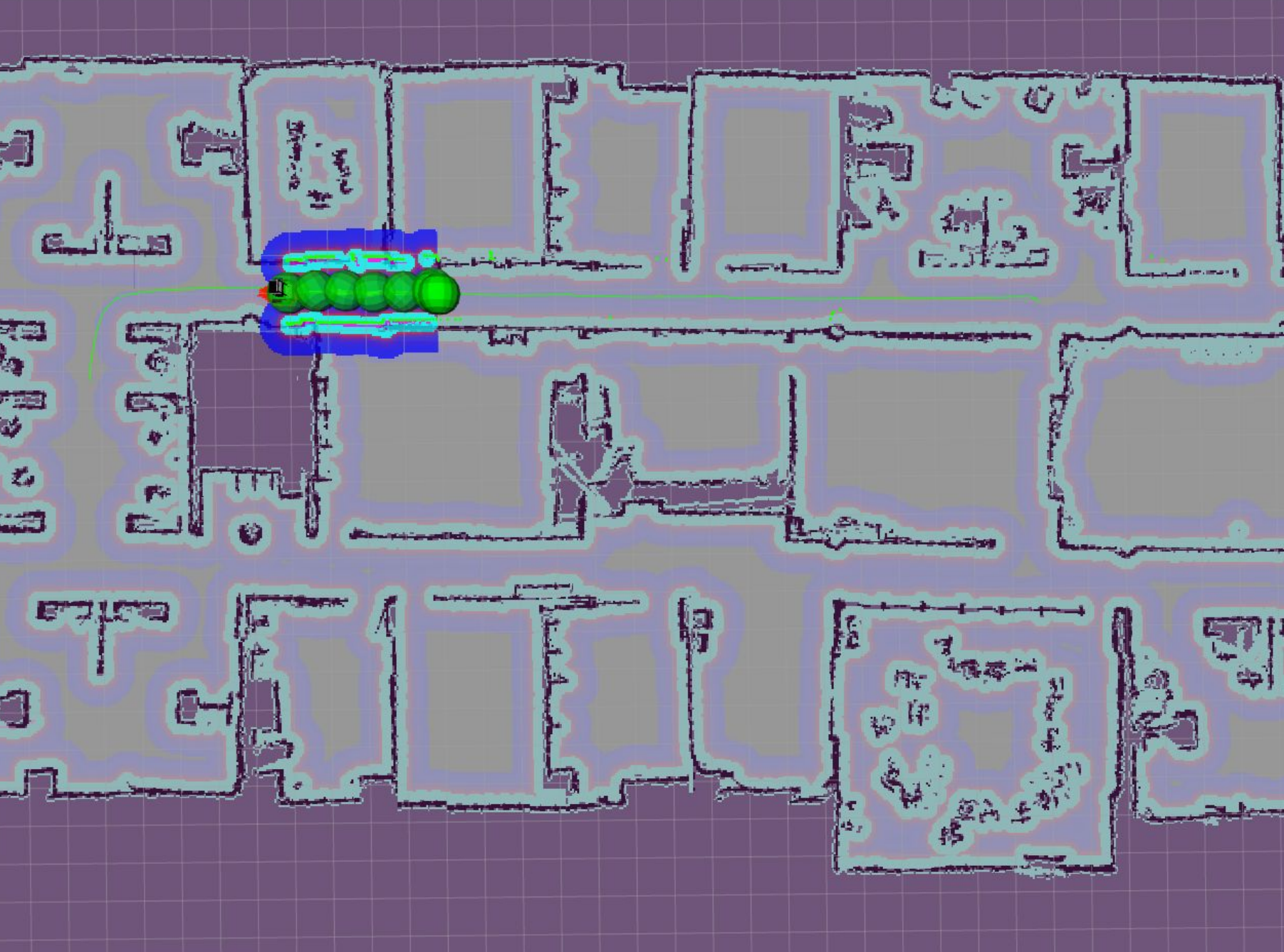}
  	\caption{Robot navigating in real environment}	
    	\label{fig:rviz}
  \end{subfigure}
  \caption{Our experiments use Gazebo to simulate an office environment and a service robot operating in this environment.}
  \label{fig:experiment_environment}
\end{figure*}

\section{Experiments}

Our mobile service robot operates autonomously as office assistants in XXX building of YYY University (citation removed for blind review). The current system performs a variety of high-level service tasks, such as finding people, answering questions, and delivering objects, which can be requested in any order during execution. While performing these tasks, navigating long distances is one of the most critical parts for the robot to achieve its goal. The dynamic characteristics of the environment make it quite challenging to achieve optimal behavior for navigation tasks because navigation costs cannot be statically determined by the distance between the start and goal locations. For example, at certain points of the day, such as when classes break, some corridors can become crowded and impossible for the robot to travel through. Furthermore, for the robot to enter rooms, it has to ask for help to open the door, while the availability of humans and the responsiveness of them heavily depend on the location and time. In this case, relying on TMP is not sufficient for the robot to behave optimally --- it needs to constantly change its behavior to adapt to the dynamic environment. Since the robot continuously operates in the same environment, it also calls for generalization from task to task: the information the robot learns about the environment from performing one task can be reused for another task, leading to enhanced long-term autonomy.

Since collecting experience on a real robot is expensive, especially when evaluating algorithms that have slower learning rates, we evaluate our approach using a simulated robot (shown in Fig.~\ref{fig:bwibot}) and an office environment (shown in Fig.~\ref{fig:gazebo}) in Gazebo. The simulation is created to closely match the real robot platform (shown in Fig.~\ref{fig:bwibot_real}) and the environment it operates in (shown in Fig.~\ref{fig:rviz}).
We compare the performance of the proposed TMP-RL algorithm (Algorithm~\ref{algexec2}) with PETLON \cite{lo2018petlon}, a TMP algorithm, and PEORL \cite{yang:peorl:2018}, a TP-RL approach which iteratively generates symbolic task plans and performs reinforcement learning during execution. We measure the actual reward the robot receives in each episode by executing the plan generated by each algorithm, and compare the learning curves over some number of episodes. Our hypothesis is that the TMP-RL algorithm outperforms the TMP-only approach when real action rewards are unexpected, and it outperforms the TP-RL approach with higher quality exploration and faster adaptation to domain changes.

\subsection{System Implementation}
The existing software on the platform is modified to implement the task-motion planning approaches.
Autonomous navigation of the robot is built on the Robot Operating System (ROS) \cite{quigley2009ros} navigation stack. A static map of the environment is pre-built with the robot's LIDAR. Path planning involves a global planner to generate path between points using Dijkstra’s algorithm, and a local planner to compute velocity commands and avoid obstacles. The map is further annotated to have regions for rooms and locations for doors and other landmarks. The connectivity of regions and the Euclidean distances between locations are also computed and stored in a central knowledge base. Two regions are connected if a path exists between them without crossing a door. To open a door in the real world, the robot requests assistance from a person nearby by speaking or displaying the request on its screen. In the Gazebo environment, doors are simulated by red obstacles shown in Figure~\ref{fig:experiment_environment}, and the \i{open\_door} action is simulated by teleporting the obstacle away after an uncertain amount of time.

We use the incremental solving mode of the answer set solver \text{Clingo} 5.3\footnote{https://github.com/potassco/clingo/releases} for generating symbolic actions. Our task planning domain models navigation actions ($\i{approach}$, $\i{open\_door}$, $\i{go\_through}$), as well as non-navigation actions (such as $\i{pick\_up}$, $\i{put\_down}$, \\ $\i{find\_person}$). In this experiment, only navigation actions need to be planned. In TMP and TMP-RL implementations, the global planner is called to generate a trajectory for each navigation action, and the motion costs are estimated by the sum of distances between waypoints on the trajectory.

We introduce our action encodings based on the notations in Section~\ref{sec:symbolic}. The following action rules encode the approach\_door action. The effects of approaching a door are: the robot will be near the door, facing the door, and in a region that is connected to the robot's current region. A door cannot be approached if the region on both sides of the door are not connected to the robot's current region.
$$
\i{approach}(D)~\causes~\i{near}(D)
$$
$$
\i{approach}(D)~\causes~\i{facing}(D)
$$
$$
\i{approach}(D)~\causes~\i{in}(R_1)~\iif~\i{has}(R_1,D),~\i{in}(R_2),~\i{connected}(R_2,R_1)
$$
$$
\begin{aligned}
\nonex~\i{approach}(D)~\iif~\sim\i{connected}(R1,R2),~\i{has}(R2,D),\\ 
~\sim\i{connected}(R1,R3),~\i{has}(R3,D)
\end{aligned}
$$

The action $\i{open\_door}$ makes a door open, and the robot must be facing the door to execute this action. This effect and precondition are formulated by the following rules.
$$
\i{open\_door}(D)~\causes~\i{open}(D)
$$
$$
\nonex~\i{open\_door}(D)~\iif~\sim\i{facing}(D)
$$

The action $\i{go\_through}$ causes the robot to be inside the room on the other side of the door. The preconditions are the robot must be facing the door and the door must be open.
$$
\i{go\_through}(D)~\causes~\i{in}(R_1)~\iif~\i{has}(R_1,D),~\i{has}(R_2,D),~\i{in}(R_2)
$$
$$
\nonex~\i{go\_through}(D)~\iif~\sim\i{facing}(D)
$$
$$
\nonex~\i{go\_through}(D)~\iif~\sim\i{open}(D)
$$

The plan quality constraint introduced in Section~\ref{sec:domain} are implemented for T(M)P-RL algorithms. Learning of the $\rho$ values is implemented using (\ref{riter1}) with learning rates $\alpha=0.1, \beta=0.5$. Our state representation consists of the $\i{near}(D)$ and $\i{in}(R)$ predicates that are true at the time. In the TMP and TMP-RL settings, the default $\rho$-values of $\i{approach}$ actions are implemented as the Euclidean distance between the target location and the landmark that the robot is currently near. The default value of an $\i{open\_door}$ action is -3, as shown in the following rule. 
$$
\default~\rho(s,\i{open\_door}(D))=-3
$$

All other state-action pairs that do not have an evaluated or default~$\rho$ value are assumed to have reward of -1.

\begin{figure*}[!hbt]
\begin{subfigure}{0.68\columnwidth}
    \includegraphics[width=\columnwidth]{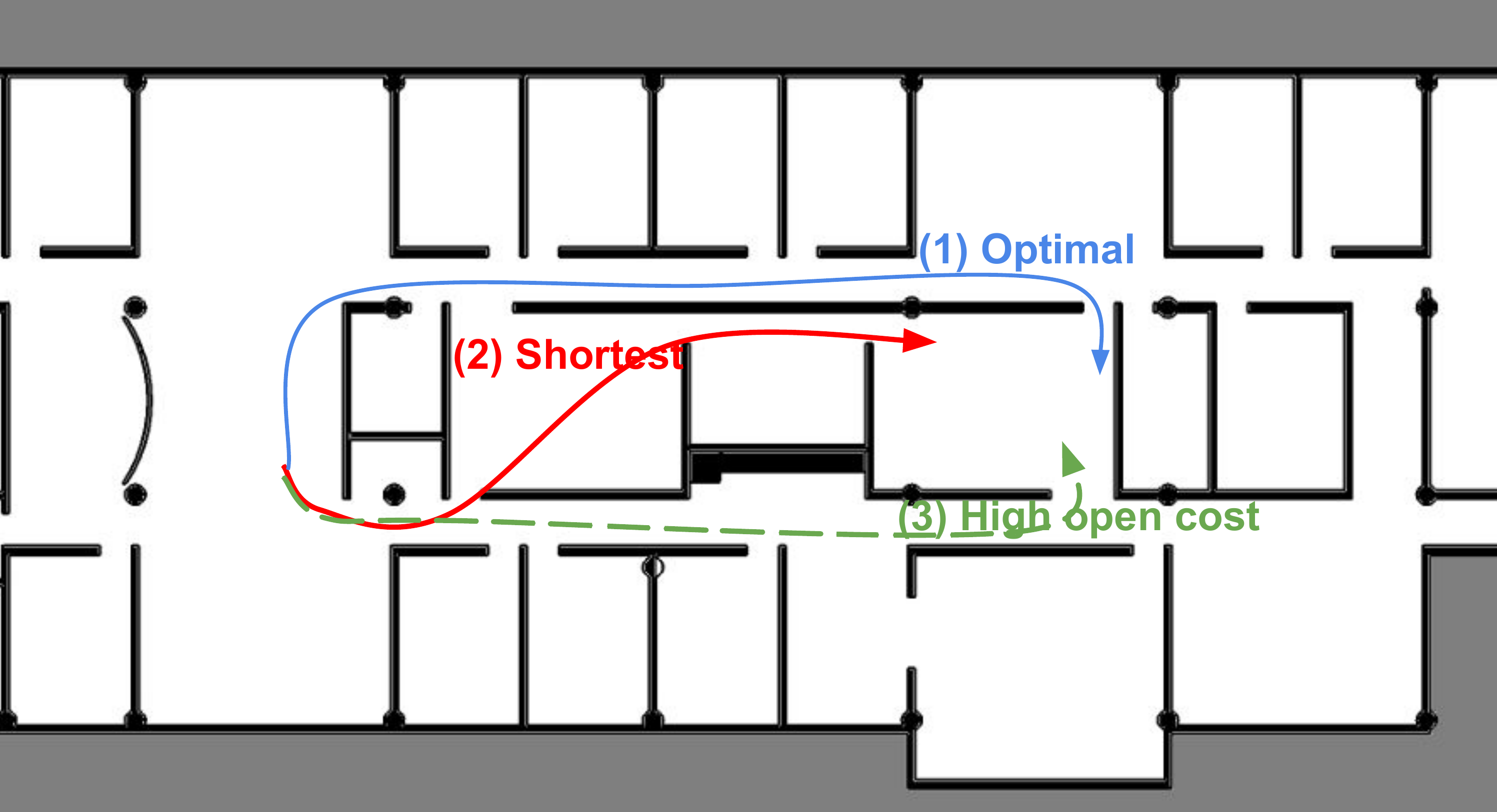}
    \caption{Three competitive plans for the task.}
    \label{fig:map}
\end{subfigure}
\quad
\begin{subfigure}{0.29\textwidth}
    \includegraphics[scale=0.3]{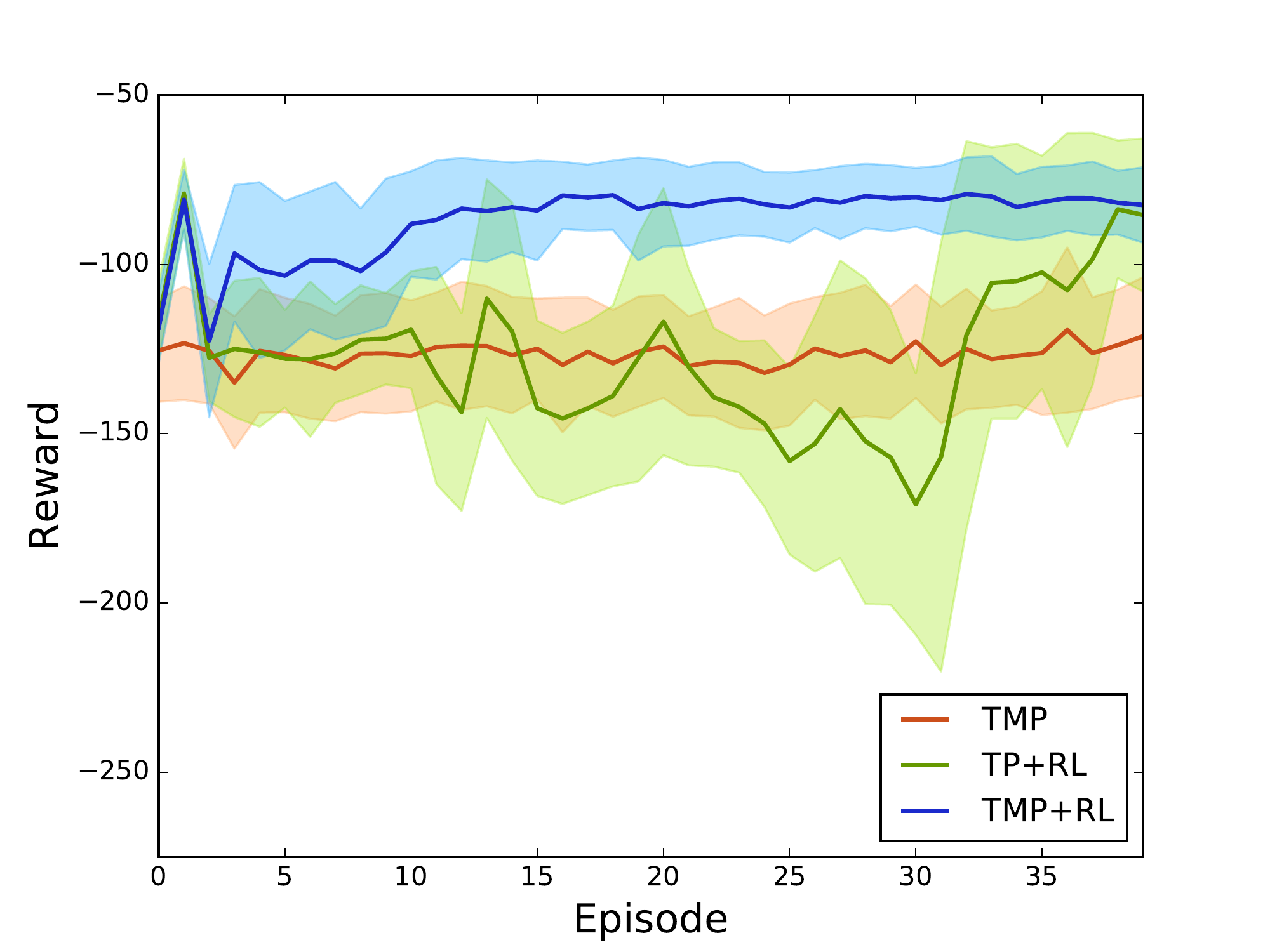}
    \caption{Learning curves.}
    \label{fig:plot}
\end{subfigure}
\quad
\begin{subfigure}{0.28\textwidth}
    \includegraphics[scale=0.3]{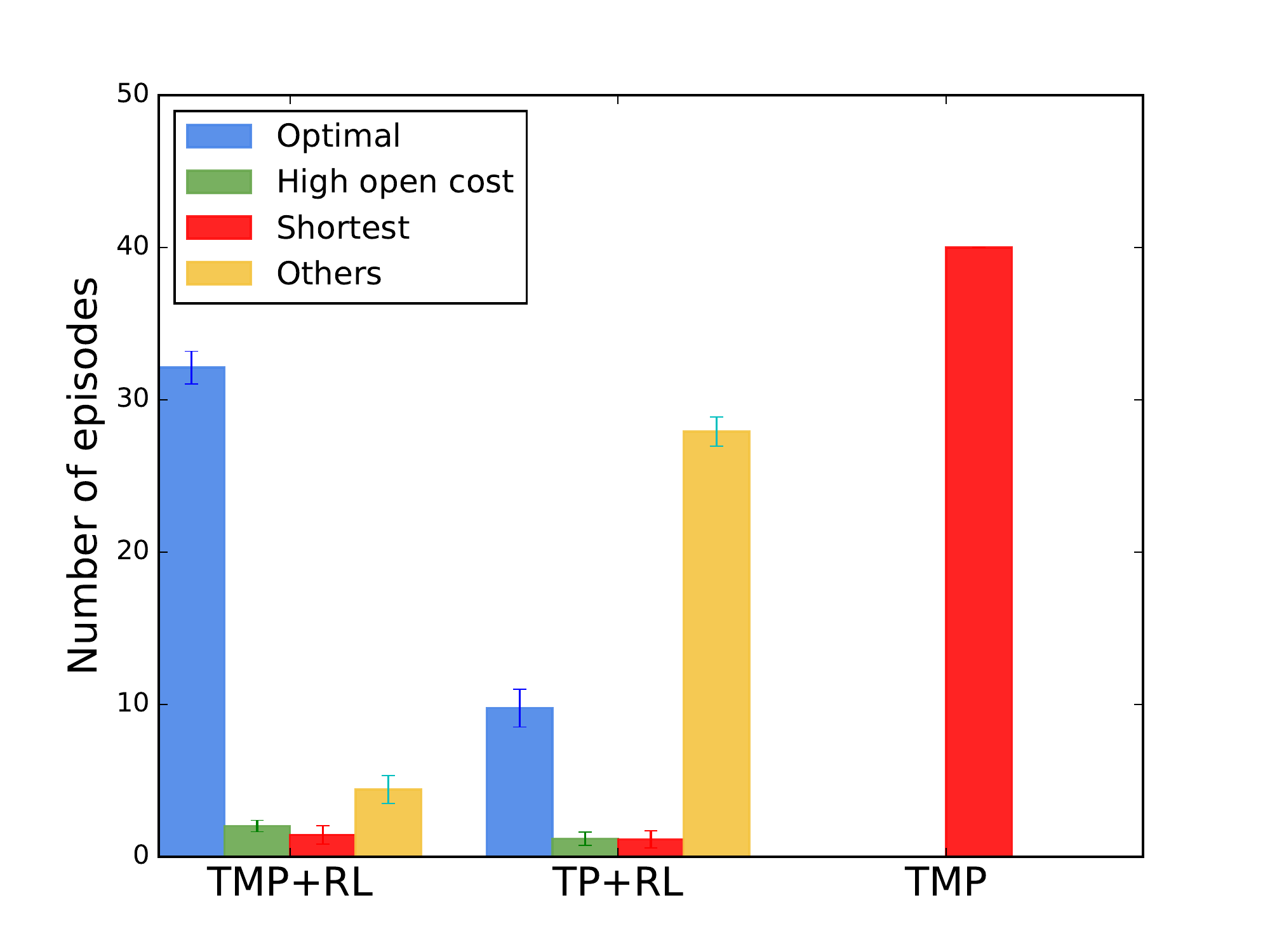}
    \caption{Distribution of plan executions.}
    \label{fig:episodes}
\end{subfigure}
\caption{Comparison of TMP, TP-RL, TMP-RL in one task.}
\label{fig:experimentresult}
\end{figure*}

\begin{table*}
  \begin{tabular}{ c | c | c | c }
    \hline
    Plan & Task Plan Length & Motion Plan Length & Average Execution Time in the Real World\\  \hline
    (1) & 3 & 60.8 & 80.6 \\ \hline
    (2) & 9 & 45.5 & 126.9\\ \hline
    (3) & 3 & 53.1 & 116.6 \\ \hline
    \hline
  \end{tabular}
  
 \caption{Plan costs at different levels of abstraction.}
 \label{tab:costs}
\end{table*}

\begin{figure*}[!hbt]
\begin{subfigure}{0.8\columnwidth}
    \includegraphics[scale=0.16]{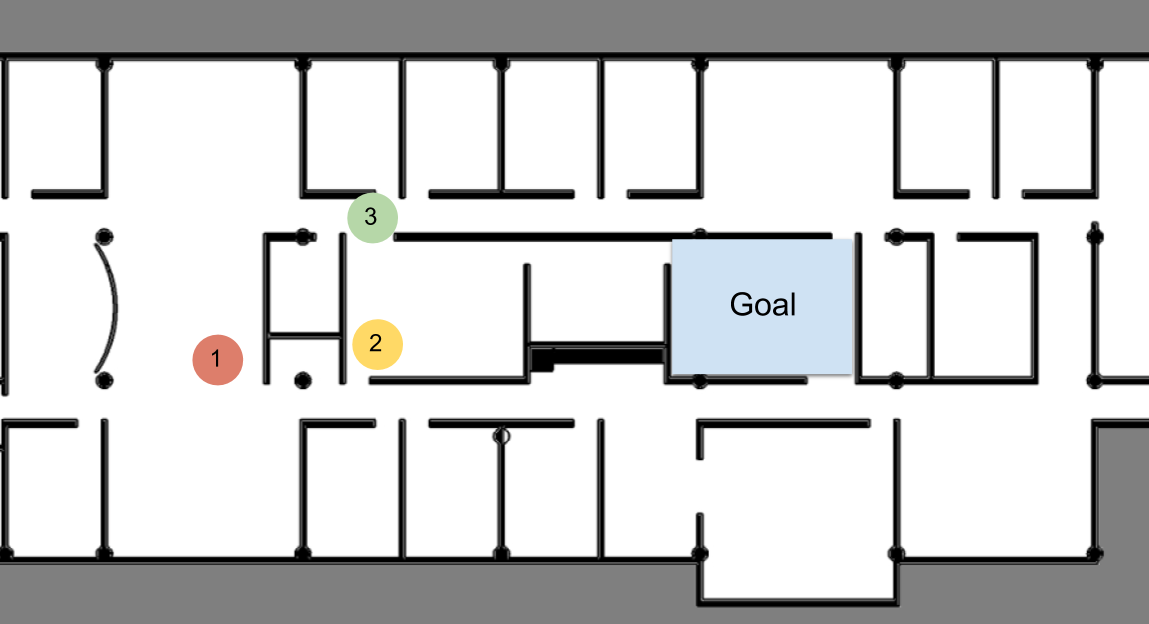}
    \caption{Extended experiment with three different initial states.}
    \label{fig:multi_task}
\end{subfigure}
\quad
\quad
\quad
\begin{subfigure}{0.4\textwidth}
    \includegraphics[scale=0.3]{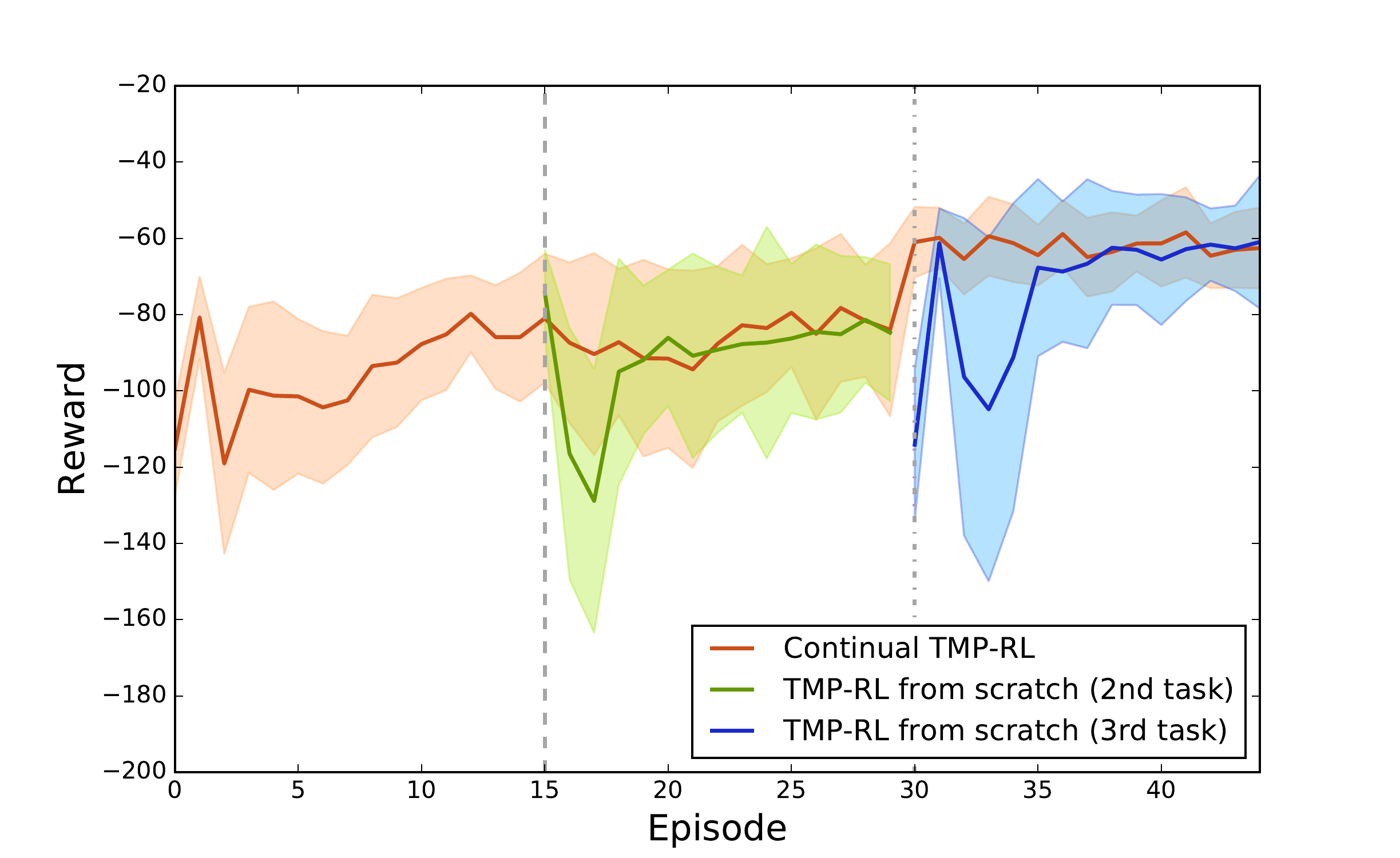}
    \caption{Learning curves of TMP-RL with continuous transfer vs. exploration from scratch.}
    \label{fig:multi_task_results}
\end{subfigure}

\caption{Experiment results for multiple task learning}
\label{fig:extendedresult}
\end{figure*}

\subsection{Experiment Design}
In this experiment, the robot's task is to go to a room where its service is requested. The robot starts near a landmark in an open space and the robot's end position can be anywhere in the target room. The reward is defined as negative of the execution time. The room has three initially closed doors that are available for entrance. The task planner determines which entrance the robot will use. 

Figure~\ref{fig:map} shows the experiment set-up and three competitive task plans. For instance, plan (1) uses the top door along the blue path and has 3 symbolic actions: $\i{approach}(top\_door)$, $\i{open\_door}(top\_door)$, $\i{go\_through}(top\_door)$. Table~\ref{tab:costs} shows the task plan length, motion plan length, and average execution time of the three competitive plans in Figure~\ref{fig:map}. Among the three plans, plan (2) features the shortest navigation distance, but it takes 9 actions and requires crossing 3 doors. Plan (3), which goes through the bottom door on the map, has 3 actions and the second shortest navigation distance. In this environment, opening door takes longer than what the robot expects, and the bottom door has a particularly high cost to open. In this simulation, the duration of executing an $\i{open\_door}$ action is sampled from a normal distribution with standard deviation equals to 10 seconds. Opening the bottom door takes 60 seconds on average, while the mean open time is 20 seconds for other doors. Therefore, plan (1) which uses the top door has the lowest expected execution time. This example shows one situation where all three levels of capability are required to efficiently find the optimal real-world plan.

Since plans that have larger number of steps might have higher cumulative reward, the planner has to search beyond the shortest plan. With 30 regions and 12 doors in the domain, many other feasible plans may be generated by task planner, and some plans involve significant detour such as approaching a location on the other side of the map before going to the target room. Therefore, generating feasible, low-cost plans efficiently and quickly adapting to domain uncertainties is quite challenging in this domain.

\subsection{Results}
\subsubsection{Evaluation of TMP, TP-RL, TMP-RL}
We first compare the performance of the three algorithms (TMP, TP-RL, TMP-RL) in this task by executing their plans in the simulator. For every approach we conducted 50 runs with 40 episodes in each run. The variability among the trials are caused by noisy action costs of navigation and opening doors. In the same episode, T(M)P-RL may generate different plans depending on experiences in previous episodes.

Figure~\ref{fig:plot} plots the learning curves for reward received in 40 episodes, averaged over 50 runs with the shaded regions representing one standard deviation from the mean. Equipped with the reinforcement learner to refine their plans, both TMP-RL and TP-RL approaches converge to the practically optimal plan, but TMP-RL converges significantly faster. The reason is that TMP-RL uses the motion planner to evaluate task plan, learns $\rho$-values and iteratively refines the task plan before executing it. Consequently, it has good value estimates for navigation actions that the robot has not experienced before and leads to a jump-start for learning in the environment. The constraint on plan quality ensures that TMP-RL makes steady improvements after the first two episodes, while TP-RL, without a motion planner to evaluate and learn $\rho$-values of the actions, learns everything from executing the plans in the environment. Consequently, TP-RL has to explore more plans before converging. The TMP approach executes the plan with the shortest navigation distance in every episode (Plan 2), and the variance in reward is only due to the noise in execution time of opening the three doors. Although it generates plan that has low cost from navigation perspective, it cannot learn the difficulty of going through the door, and leads to suboptimal behavior.

Figure~\ref{fig:plot} also shows that both TMP and TMP-RL have smaller variance during execution, due to the fact that the motion planner provides a baseline evaluation on the quality of the plan. TP-RL also has low variance in the first two episodes, because the task planner first selects the plans with the smallest number of actions (plan (1) and (3) in Figure~\ref{fig:map}), but much higher variance afterwards, indicating that many task plans that are logically valid but significantly worse in quality are executed and evaluated directly in the environment, which is expensive and potentially dangerous for real robots. 

Figure~\ref{fig:episodes} shows for each approach, the average number of episodes that the three competitive plans and other feasible plans are executed. Compared to TMP-RL, TP-RL spends a lot more exploration on less competitive plans. Although it can eventually adapt to domain uncertainties, the learning is slow and costly. TMP always executes the same plan with no exploration. These observations validate our hypothesis that the proposed TMP-RL approach enables more robust planning and faster adaptation in dynamic domains by closely combining task planning, motion planning, and learning.

In terms of planning time, each call of the answer set solver is timed-out after 5 seconds for all three approaches. The ``anytime'' property of PETLON allows it to find plans of good quality given early termination. This property also holds for Algorithm~\ref{algexec} and ~\ref{algexec2} since they incrementally tighten the bound of plan quality. When the solver fails to find a better plan within the time-out, the algorithms return the current best plan.

\subsubsection{Evaluation of TMP-RL in Multiple Tasks}
In long-term deployments, the robot can be asked to achieve the same end goal from different starting positions. For example, the robot may start in the mail room and deliver mail to an office in the morning, and deliver coffee from the kitchen to the same office in the afternoon. Since the initial states are different, the task planner and motion planner have to solve them as different problems, but TMP-RL can leverage the learned $rho$-values to speed up exploration in later tasks. In order to demonstrate the TMP-RL framework's ability to generalize learned values to different scenarios, we extend the previous experiment with two more tasks, each with a different starting position of the robot (shown in Figure~\ref{fig:multi_task}). 

In this scenario, we compare continuously running TMP-RL for all three tasks against using TMP-RL to learn the second and the third tasks from scratch. In the former setting, the robot explores the first task for 15 episodes, and switches to the second position and the third position while keeping the learned values. In the latter setting, the robot starts up at episode 15 and performs the second task, or starts up at episode 30 and performs the third task.
Figure~\ref{fig:multi_task_results} presents the learning curves averaged over 40 runs in these three settings, showing that learning the first task leads to faster, lower variance learning in the later tasks, in comparison with learning from scratch, indicating that the learned values can be transferred to accelerate learning other tasks.

\section{Conclusion}
We propose a novel TMP-RL framework integrating task-motion planning (TMP) and reinforcement learning (RL) for robot decision making. The framework mixes task planning, motion planning and reinforcement learning in a closed loop with iterative improvements on plan quality over the course of execution. We evaluate and compare TMP-RL with TMP and TP-RL approaches in a realistic simulation of a service robot domain. The experiments demonstrate that TMP-RL combines the strengths of the individual paradigms: task-motion planning generates high-quality plans without performing costly learning in the real environment, and reinforcement learning refines task-motion planning in dynamic domains and generalizes learned information to new scenarios. Therefore, this framework provides important properties for long-term deployments of robots in dynamic environments.

\bibliographystyle{ACM-Reference-Format}
\bibliography{bib}

\end{document}